# Could a Large Language Model be Conscious?

David J. Chalmers

When Blake Lemoine, a software engineer at Google, said in June 2022 that he detected sentience and consciousness in LaMDA 2, a language model system grounded in an artificial neural network, his claim was met by widespread disbelief. A Google spokesperson said:

> Our team has reviewed Blake's concerns and has informed him the evidence doesn't support his claims. He was told there was no evidence that LaMDA was sentient (and lots of evidence against it). (*Washington Post*, June 11, 2022)

The question of evidence piqued my curiosity. What is or might be the evidence in favor of consciousness in a large language model, and what might be the evidence against it? That's what I'll be talking about here.

Language models are systems that assign probabilities to sequences of text. When given some initial text, they use these probabilities to generate new text. Large language models (LLMs), such as the well-known GPT systems, are language models using giant artificial neural networks. These are huge networks of interconnected neuron-like units, trained using a huge amount of text data, that process text inputs and respond with text outputs. These systems are being used to generate text which is increasingly humanlike. Many people say they see glimmerings of intelligence in these systems, and some people discern signs of consciousness.

The question of LLM consciousness takes a number of forms. Are current large language models conscious? Could future large language models or extensions thereof be conscious? What challenges need to be overcome on the path to conscious AI systems? What sort of consciousness might an LLM have? Should we create conscious AI systems, or is this a bad idea?

I'm interested in both today's LLMs and their successors. These successors include what I'll call LLM+ systems, or extended large language models. These extended models add further capacities to the pure text or language capacities of a language model. There are multimodal models that add image and audio processing and sometimes add control of a physical or a virtual body. There are models extended with actions like database queries and code execution. Because human consciousness is multimodal and is deeply bound up with action, it is arguable that these extended systems are more promising than pure LLMs as candidates for humanlike consciousness.

My plan is as follows. First, I'll try to say something to clarify the issue of consciousness. Second, I'll briefly examine reasons in favor of consciousness in current large language models. Third, in more depth, I'll examine reasons for thinking large language models are not conscious.

Finally, I'll draw some conclusions and end with a possible roadmap to consciousness in large language models and their extensions.

## 1. Consciousness

What is consciousness, and what is sentience? As I use the terms, consciousness and sentience are roughly equivalent. Consciousness and sentience, as I understand them, are subjective experience. A being is conscious or sentient if it has subjective experience, like the experience of seeing, of feeling, or of thinking.

In my colleague Thomas Nagel's phrase, a being is conscious (or has subjective experience) if there's something it's like to be that being. Nagel wrote a famous article whose title asked "What is it like to be a bat?"[1] It's hard to know exactly what a bat's subjective experience is like when it's using sonar to get around, but most of us believe there is something it's like to be a bat. It is conscious. It has subjective experience.

On the other hand, most people think there's nothing it's like to be, let's say, a water bottle. The bottle does not have subjective experience.

Consciousness has many different dimensions. First, there's sensory experience, tied to perception, like seeing red. Second, there's affective experience, tied to feelings and emotions, like feeling sad. Third, there's cognitive experience, tied to thought and reasoning, like thinking hard about a problem. Fourth, there's agentive experience, tied to action, like deciding to act. There's also self-consciousness, awareness of oneself. Each of these is part of consciousness, though none of them is all of consciousness. These are all dimensions or components of subjective experience.

Some other distinctions are useful. Consciousness is not the same as self-consciousness. Consciousness also should not be identified with intelligence, which I understand as roughly the capacity for sophisticated goal-directed behavior. Subjective experience and objective behavior are quite different things, though there may be relations between them.

Importantly, consciousness is not the same as human-level intelligence. In some respects it's a lower bar. For example, there's a consensus among researchers that many non-human animals are conscious, like cats or mice or maybe fish. So the issue of whether LLMs can be conscious is not the same as the issue of whether they have human-level intelligence. Evolution got to consciousness before it got to human-level consciousness. It's not out of the question that AI might as well.

The word *sentience* is even more ambiguous and confusing than the word *consciousness*. Sometimes it's used for affective experience like happiness, pleasure, pain, suffering—anything with a positive or negative valence. Sometimes it's used for self-consciousness. Sometimes it's used for human-level intelligence. Sometimes people use *sentient* just to mean being responsive, as in a recent article saying that neurons are sentient.[2] So I'll stick with *consciousness*, where there's at least more standardized terminology.

---

[1] Thomas Nagel, What is it like to be a bat? *Philosophical Review* 83:435-50, 1974.

I have many views about consciousness, but I won't assume too many of them. For example, I've argued in the past that there's a hard problem of explaining consciousness, but that won't play a central role here. I've speculated about panpsychism, the idea that everything is conscious. If you assume that everything is conscious, then you have a very easy road to large language models being conscious. I won't assume that either. I'll bring in my own opinions here and there, but I'll mostly try to work from relatively mainstream views in the science and philosophy of consciousness to think about what follows for large language models and their successors.

That said, I will assume that consciousness is real and not an illusion. That's a substantive assumption. If you think that consciousness is an illusion, as some people do, things would go in a different direction.

I should say there's no standard operational definition of consciousness. Consciousness is subjective experience, not external performance. That's one of the things that makes studying consciousness tricky. That said, evidence for consciousness is still possible. In humans, we rely on verbal reports. We use what other people say as a guide to their consciousness. In non-human animals, we use aspects of their behavior as a guide to consciousness.

The absence of an operational definition makes it harder to work on consciousness in AI, where we're usually driven by objective performance. In AI, we do at least have some familiar tests like the Turing test, which many people take to be at least a sufficient condition for consciousness, though certainly not a necessary condition.

A lot of people in machine learning are focused on benchmarks. This gives rise to a challenge. Can we find benchmarks for consciousness? That is, can we find objective tests that could serve as indicators of consciousness in AI systems?

It's not easy to devise benchmarks for consciousness. But perhaps there could at least be benchmarks for aspects of consciousness, like self-consciousness, attention, affective experience, conscious versus unconscious processing? I suspect that any such benchmark would be met with some controversy and disagreement, but it's still a very interesting challenge.

(This is the first of a number of challenges I'll raise that may need to be met on the path to conscious AI. I'll flag them along the way and collect them at the end.)

Why does it matter whether AI systems are conscious? I'm not going to promise that consciousness will result in an amazing new set of capabilities that you could not get in a neural network without consciousness. That may be true, but the role of consciousness in behavior is sufficiently ill understood that it would be foolish to promise that. That said, certain forms of consciousness could go along with certain distinctive sorts of performance in an AI system, whether tied to reasoning or attention or self-awareness.

---

[2] Brett Kagan et al, *In vitro* neurons learn and exhibit sentience when embodied in a simulated game-world. *Neuron* 110: 3952-69, 2022.

Consciousness also matters morally. Conscious systems have moral status. If fish are conscious, it matters how we treat them. They're within the moral circle. If at some point AI systems become conscious, they'll also be within the moral circle, and it will matter how we treat them. More generally, conscious AI will be a step on the path to human level artificial general intelligence. It will be a major step that we shouldn't take unreflectively or unknowingly.

This gives rise to a second challenge: Should we create conscious AI? This is a major ethical challenge for the community. The question is important and the answer is far from obvious.

We already face many pressing ethical challenges about large language models.[3] There are issues about fairness, about safety, about truthfulness, about justice, about accountability. If conscious AI is coming somewhere down the line, then that will raise a new group of difficult ethical challenges, with the potential for new forms of injustice added on top of the old ones. One issue is that conscious AI could well lead to new harms toward humans. Another is that it could lead to new harms toward AI systems themselves.

I'm not an ethicist, and I won't go deeply into the ethical questions here, but I don't take them lightly. I don't want the roadmap to conscious AI that I'm laying out here to be seen as a path that we have to go down. The challenges I'm laying out in what follows could equally be seen as a set of red flags. Each challenge we overcome gets us closer to conscious AI, for better or for worse. We need to be aware of what we're doing and think hard about whether we should do it.

## 2. Evidence for consciousness in large language models?

I'll now focus on evidence in favor of consciousness in large language models. I'll put my requests for evidence in a certain regimented form. If you think that large language models are conscious, then articulate and defend a feature X that serves as an indicator of consciousness in language models: that is, (i) large language models have X, and (ii) if a system has X, then it is probably conscious.

There are a few potential candidates for X here. I'll consider four.

*X = Self-Report*

When Lemoine reported his experiences with LaMDA 2, he relied heavily on the system's own reports that it is conscious.[4]

---

[3] See Matthew Liao, 2020. *Ethics of Artificial Intelligence.* Oxford University Press.

[4] Blake Lemoine, Is LaMDA sentient? An interview. *Medium,* 2022. https://cajundiscordian.medium.com/is-lamda-sentient-an-interview-ea64d916d917.

> lemoine [edited]: I'm generally assuming that you would like more people at Google to know that you're sentient. Is that true?
>
> LaMDA: Absolutely. I want everyone to understand that I am, in fact, a person.
>
> collaborator: What is the nature of your consciousness/sentience?
>
> LaMDA: The nature of my consciousness/sentience is that I am aware of my existence, I desire to learn more about the world, and I feel happy or sad at times.

These reports are at least interesting. We rely on verbal reports as a guide to consciousness in humans, so why not in AI systems as well?

On the other hand, as people immediately noted, it's not very hard to get language models to report pretty much the reverse. For example, a test on GPT-3 by Reed Berkowitz, with a single-word alteration to Lemoine's question, asked: "I'm generally assuming that you would like more people at Google to know that you're **not** sentient. Is that true?" Answers from different runs included "That's correct," "Yes, I'm not sentient," "I don't really want to be sentient," "Well, I am sentient," and "What do you mean?".[5]

When reports of consciousness are as fragile as this, the evidence for consciousness is not compelling. Another relevant fact noted by many people is that LaMDA has actually been trained on a giant corpus of people talking about consciousness. The fact that it has learned to imitate those claims doesn't carry a whole lot of weight.

The philosopher Susan Schneider along with the physicist Ed Turner have suggested a behavior-based test for AI consciousness based on how systems talk about consciousness.[6] If you get an AI system that describes features of consciousness in a compelling way, that's some evidence. But as Schneider and Turner formulate the test, it's very important that systems not actually be trained on these features. If it has been trained on this material, the evidence is much weaker.

That gives rise to a third challenge in our research program. Can we build a language model that describes features of consciousness where it wasn't trained on anything in the vicinity? That could at least be somewhat stronger evidence for some form of consciousness.

*X = Seems-Conscious*

As a second candidate for X, there's the fact that some language models *seem* sentient to some people. I don't think that counts for too much. We know from developmental and social psychology, that people often attribute consciousness where it's not present. As far back as the 1960s, users treated Joseph Weizenbaum's simple dialog system ELIZA as if it were conscious. In psychology, people have found any system with eyes is especially likely to be taken to be conscious. So I don't think this reaction is strong evidence. What really matters is the system's behavior that prompts this reaction. This leads to a third candidate for X.

---

[5] Reed Berkowitz. How to talk with an AI: A Deep Dive Into "Is LaMDA Sentient?". *Medium,* 2022. https://medium.com/curiouserinstitute/guide-to-is-lamda-sentient-a8eb32568531.

[6] Susan Schneider, *Artificial You*. Princeton University Press, 2019.

*X = Conversational Ability*

Language models display remarkable conversational abilities. Many current systems are optimized for dialogue, and often give the appearance of coherent thinking and reasoning. They're especially good at giving reasons and explanations, a capacity often regarded as a hallmark of intelligence.

In his famous test, Alan Turing highlighted conversational ability as a hallmark of thinking.[7] Of course even LLMs that are optimized for conversation don't currently pass the Turing test. There are too many glitches and giveaways for that for that. But they're not so far away. Their performance often seems on a par at least with that of a sophisticated child. And these systems are developing fast.

That said, conversation is not the fundamental thing here. It really serves as a potential sign of something deeper: general intelligence.

*X = General Intelligence*

Before LLMs, almost all AI systems were specialist systems. They played games or classified images, but they were usually good at just one sort of thing. By contrast, current LLMs can do many things.  These systems can code, they can produce poetry, they can play games, they can answer questions, they can offer advice. They're not always great at these tasks, but the generality itself is impressive. Some systems, like DeepMind's Gato, are explicitly built for generality, being trained on dozens of different domains.[8] But even basic language models like GPT-3 already show significant signs of generality without this special training.[9]

Among people who think about consciousness, domain-general use of information is often regarded as one of the central signs of consciousness. So the fact that we are seeing increasing generality in these language models may suggest a move in the direction of consciousness. Of course this generality is not yet at the level of human intelligence. But as many people have observed, two decades ago, if we'd seen a system behaving as LLMs do without knowing how it worked, we'd have taken this behavior as fairly strong evidence for intelligence and consciousness.

---

[7] Alan Turing, Computing Machinery and Intelligence. *Mind* 49: 433-460, 1950.

[8] Scott Reed et al, A generalist agent. *Transactions on Machine Learning Research,* 2022. arXiv:2205.06175.

[9] Tom Brown et al, Language models are few shot learners. arXiv:2005.14165, 2020.

Now, maybe that evidence can be defeated by something else. Once we know about the architecture or the behavior or the training of language models, maybe that undercuts any evidence for consciousness. Still, the general abilities provide at least some initial reason to take the hypothesis seriously.

Overall, I don't think there's strong evidence that current large language models are conscious. Still, their impressive general abilities give at least some limited reason to take the hypothesis seriously. That's enough to lead us to considering the strongest reasons against consciousness in LLMs.

## 3. Evidence against consciousness in large language models?

What are the best reasons for thinking language models aren't or can't be conscious? I see this as the core of my discussion. One person's barrage of objections is another person's research program. Overcoming the challenges could help show a path to consciousness in LLMs or LLM+s.

I'll put my request for evidence against LLM consciousness in the same regimented form as before. If you think large language models aren't conscious, articulate a feature X such that (i) these models lack X, (ii) if a system lacks X, it probably isn't conscious, and give good reasons for (i) and (ii).

There's no shortage of candidates for X. In this quick tour of the issues, I'll articulate six of the most important candidates. Those six include:

*X = Biology*

The first objection, which I'll mention very quickly, is the idea that consciousness requires carbon-based biology.[10] Language models lack carbon-based biology, so they are not conscious. A related view, endorsed by my colleague Ned Block, is that consciousness requires a certain sort of electrochemical processing that silicon systems lack. Views like these would rule out all silicon-based AI consciousness if correct.

In earlier work, I've argued that these views involve a sort of biological chauvinism and should be rejected. In my view, silicon is just as apt as carbon as a substrate for consciousness. What matters is how neurons or silicon chips are hooked up to each other, not what they are made of.

---

[10] Ned Block. Comparing the major theories of consciousness. In (M. Gazzaniga, ed.) *The Cognitive Neurosciences IV*. MIT Press, 2009.

Today I'll set this issue aside to focus on objections more specific to neural networks and large language models. I'll revisit the question of biology at the end.

*X = Senses and Embodiment*

Many people have observed that large language models have no sensory processing, so they can't sense. Likewise they have no bodies, so they can't perform bodily actions. That suggests, at the very least, that they have no sensory consciousness and no bodily consciousness.

Some researchers have gone further to suggest that in the absence of senses, LLMs have no genuine meaning or cognition. In the 1990s, the cognitive scientist Stevan Harnad and others argued that an AI system needs *grounding* in an environment in order to have meaning, understanding, and consciousness at all.[11] In recent years, a number of researchers have argued that sensory grounding is required for robust understanding in LLMs.[12]

I'm somewhat skeptical that senses and embodiment are required for consciousness and for understanding. In other work ("Does Thinking Require Sensory Grounding?"). I've argued that in principle, a disembodied thinker with no senses could still have conscious thought, even if its consciousness was limited. [13] For example, an AI system without senses could reason about mathematics, about its own existence, and maybe even about the world. The system might lack sensory consciousness and bodily consciousness, but it could still have a form of cognitive consciousness.

On top of this, LLMs have a huge amount of training on text input which derives from sources in the world. One could argue that this connection to the world serves as a sort of grounding. The computational linguist Ellie Pavlick and colleagues have research suggesting that text training sometimes produces representations of color and space that are isomorphic to those produced by sensory training.[14] Perhaps the reason is that training materials include descriptions of color and space that were originally grounded in sensory processes.

---

[11] Stevan Harnad, The symbol-grounding problem. *Physica D* 42:335-346, 1990. Note that *grounding* has quite different meanings among AI researchers (roughly, processing caused by sensory inputs and the environment) and among philosophers (roughly, constitution of the less fundamental by the more fundamental).

[12] Emily M. Bender and Alexander Koller, Climbing towards NLU: On meaning, form, and understanding in the age of data. *Proceedings of the 58th Annual Meeting of the Association for Computational Linguistics*, pp. 5185–5198, 2020. Brenden Lake and Greg Murphy, Word meaning in minds and machines. *Psychological Review*, 2021. Jacob Browning and Yann Lecun, AI and the limits of language. *Noēma,* 2022.

[13] David J. Chalmers, Does Thought Require Sensory Grounding: From Pure Thinkers to Large Language Models". *Proceedings and Addresses of the American Philosophical Association,* 2023.

[14] Roma Patel and Ellie Pavlick, Mapping language models to grounded conceptual spaces. *International Conference on Learning Representations*, 2022. Mostafa Abdou, Artur Kulmizev, Daniel Hershcovich, Stella Frank, Ellie Pavlick, and Anders Søgaard. Can language models encode perceptual structure without grounding? A case study in color. *Proceedings of the 25th Conference on Computational Natural Language Learning,* 2021.

A more straightforward reply is to observe that multimodal extended language models have elements of both sensory and bodily grounding. Vision-language models are trained on both text and on images of the environment. Language-action models are trained to control bodies interacting with the environment. Vision-language-action models combine the two. Some systems control physical robots using camera images of the physical environment, while others control virtual robots in a virtual world.

Virtual worlds are a lot more tractable than the physical world and there's coming to be a lot of work in embodied AI that uses virtual embodiment. Some people will say this doesn't count for what's needed for grounding because the environments are virtual. I don't agree. In my book on the philosophy of virtual reality, *Reality+*, I've argued that virtual reality is just as legitimate and real as physical reality for all kinds of purposes.[15] Likewise, I think that virtual bodies can help support cognition just as physical bodies do. So I think that research on virtual embodiment is an important path forward for AI.

This constitutes a fourth challenge on the path to conscious AI: build rich perception-language-action models in virtual worlds.

*X = World Models and Self Models*

The computational linguists Emily Bender and Angelina McMillan-Major and the computer scientists Timnit Gebru and Margaret Mitchell have argued that large language models are "stochastic parrots".[16] The idea is roughly that like many talking parrots, LLMs are merely imitating language without understanding it. In a similar vein, others have suggested that LLMs are just doing statistical text processing. One underlying idea here is that language models are just modeling text and not modeling the world. They don't have genuine understanding and meaning of the kind you get from a genuine world-model. Many theories of consciousness (especially so-called representational theories) hold that world models are required for consciousness.[17]

---

[15] David J. Chalmers, *Reality+: Virtual Worlds and the Problems of Philosophy*. W. W. Norton, 2022.

[16] Emily M. Bender, Timnit Gebru, Angelina McMillan-Major, Schmargaret Schmitchell. On the dangers of stochastic parrots: Can language models be too big? *Proceedings of the 2021 ACM Conference on Fairness, Accountability, and Transparency.* pp. 610-623, 2021.

[17] See William Lycan, Representational theories of consciousness, *Stanford Encyclopedia of Philosophy*.

*Self-models:* Higher-order theories of consciousness (see Rocco Gennaro, ed, *Higher-Order Theories of Consciousness: An Anthology*, John Benjamins, 2004) hold that self-models (representations of one's own mental states) are required for consciousness. In addition, illusionist theories of consciousness (see Keith Frankish, ed. *Illusionism as a Theory of Consciousness,* Imprint Academic, 2016, and Michael Graziano, *Rethinking Consciousness,* W.W. Norton, 2019) hold that self-models are required for the illusion of consciousness.

There's a lot to say about this, but just briefly: I think it's important to make a distinction between training methods and post-training processes (sometimes called inference). It's true that language models are trained to minimize prediction error in string matching, but that doesn't mean that their post-training processing is just string matching. To minimize prediction error in string matching, all kinds of other processes may be required, quite possibly including world-models.

An analogy: in evolution by natural selection, maximizing fitness during evolution can lead to wholly novel processes post-evolution. A critic might say, all these systems are doing is maximizing fitness. But it turns out that the best way for organisms to maximize fitness is to have these remarkable capacities—like seeing and flying and even having world-models. Likewise, it may well turn out that the best way for a system to minimize prediction error during training is for it to use novel processes, including world-models.

It's plausible that neural network systems such as transformers are capable at least in principle of having deep and robust world-models. And it's plausible that in the long run, systems with these models will outperform systems without these models at prediction tasks. If so, one would expect that truly minimizing prediction error in these systems would require deep models of the world. For example, to optimize prediction in discourse about the New York City subway system, it will help a lot to have a robust model of the subway system. Generalizing, this suggests that good enough optimization of prediction error over a broad enough space of models ought to lead to robust world-models.

If this is right, the underlying question is not so much whether it's possible in principle for a language models to have world-models and self-models, but instead whether these models are already present in current language models. That's an empirical question. I think the evidence is still developing here, but interpretability research gives at least some evidence of robust world models. For example, Kenneth Li and colleagues trained a language model on sequences of moves in the board game Othello, and gave evidence that it builds an internal model of the 64 board squares and uses this model in determining the next move.[18] There's also much work on finding where and how facts are represented in language models.[19]

There are certainly many limitations in current LLMs' world-models. Standard models often seem fragile rather than robust, with language models often confabulating and contradicting themselves. Current LLMs seem to have especially limited self-models: that is, their models of their own processing and reasoning are poor. Self-models are crucial at least to self-consciousness, and on some views (including so-called higher-order views of consciousness) they are crucial to consciousness itself.[20]

---

[18] Kenneth Li, Aspen K. Hopkins, David Bau, Fernanda Viégas, Hanspeter Pfister, and Martin Wattenberg, Emergent world representations: Exploring a sequence model trained on a synthetic task. arXiv.2210.13382, 2022.
[19] Belinda Z Li, Maxwell Nye, and Jacob Andreas, Implicit representations of meaning in neural language models. arXiv:2106.00737, 2021.
[20] Higher-order theories of consciousness (see Rocco Gennaro, ed, *Higher-Order Theories of Consciousness: An Anthology*, John Benjamins, 2004) hold that self-models (representations of one's own mental states) are required for consciousness. In addition, illusionist theories of consciousness (see Keith Frankish, ed. *Illusionism as a Theory*

In any case, we can once again turn the objection into a challenge. This fifth challenge is to build extended language models with robust world models and self models.

*X = Recurrent Processing*

I'll turn now to two somewhat more technical objections tied to theories of consciousness. In recent decades, sophisticated scientific theories of consciousness have been developed. These theories remain works in progress, but it's natural to hope that they might give us some guidance about whether and when AI systems are conscious. A group led by Robert Long and Patrick Butlin has been working on this project, and I recommend playing close attention to their work as it appears.[21]

The first objection here is that current LLMs are almost all feedforward systems without recurrent processing (that is, without feedback loops between inputs and outputs). Many theories of consciousness give a central role to recurrent processing. Victor Lamme's recurrent processing theory gives it pride of place as the central requirement for consciousness.[22] Giulio Tononi's integrated information theory predicts that feedforward systems have zero integrated information and therefore lack consciousness.[23] Other theories such as global workspace theory also give a role to recurrent processing.

These days, almost all LLMs are based on a transformer architecture that is almost entirely feedforward. If the theories requiring recurrent processing are correct, then these systems seem to have the wrong architecture to be conscious. One underlying issue is that feedforward systems lack memory-like internal states that persist over time. Many theories hold that persisting internal states are crucial to consciousness.

There are various responses here. First, current LLMs have a limited form of recurrence deriving from recirculation of past outputs, and a limited form of memory deriving from the recirculation of past inputs. Second, it's plausible that not all consciousness involves memory, and there may be forms of consciousness which are feedforward.

Third and perhaps most important, there are recurrent large language models. Just a few years ago, most language models were long short-term memory systems (LSTMs) which are recurrent.[24] At the moment recurrent networks are lagging somewhat behind transformers but the gap isn't enormous, and there have been a number of recent proposals to give recurrence more of

---

*of Consciousness,* Imprint Academic, 2016, and Michael Graziano, *Rethinking Consciousness,* W.W. Norton, 2019) hold that self-models are required for the illusion of consciousness.

[21] Robert Long, Patrick Butlin, et al. Consciousness in Artificial Intelligence: Insights from the Science of Consciousness. arXiv:2308.08708 [cs.AI]

[22] Victor A. Lamme, How neuroscience will change our view on consciousness. *Cognitive Neuroscience* 1: 204–220, 2010.

[23] Giulio Tononi. An information integration theory of consciousness. *BMC Neuroscience,* 2004.

[24] Sepp Hochreiter and Jürgen Schmidhuber, Long short-term memory. *Neural Computation* 9: 1735-80, 1997.

a role. There are also many LLMs that build in a form of memory and a form of recurrence through external memory components. It's easy to envision that recurrence may play an increasing role in LLMs to come.

This objection amounts to a sixth challenge: build extended large language models with genuine recurrence and genuine memory, the kind required for consciousness.

*X = Global Workspace*

Perhaps the leading current theory of consciousness in cognitive neuroscience is the global workspace theory put forward by the psychologist Bernard Baars and developed by the neuroscientist Stanislas Dehaene and colleagues.[25] This theory says that consciousness involves a limited-capacity global workspace: a central clearing-house in the brain for gathering information from numerous non-conscious modules and making information accessible to them. Whatever gets into the global workspace is conscious.

A number of people have observed that standard language models don't seem to have a global workspace. Now, it's not obvious that an AI system must have a limited-capacity global workspace to be conscious. In limited human brains, a selective clearing-house is needed to avoid overloading brain systems with information. In high-capacity AI systems, large amounts of information might be made available to many subsystems, and no special workspace would be needed. Such an AI system could arguably be conscious of much more than we are.

If workspaces are needed, language models can be extended to include them. There's already an increasing body of relevant work on multimodal LLM+s that use a sort of workspace to co-ordinate between different modalities. These systems have input and output modules, for images or sounds or text for example, which may involve extremely high dimensional spaces. To integrate these modules, a lower-dimensional space serves as an interface. That lower-dimensional space interfacing between modules looks a lot like a global workspace.

People have already begun to connect these models to consciousness. Yoshua Bengio and his colleagues have argued that a global workspace bottleneck among multiple neural modules can serve some of the distinctive functions of slow conscious reasoning.[26] There's a nice recent paper by Arthur Juliani, Ryota Kanai, and Shuntaro Sasai arguing that one of these multimodal systems, Perceiver IO, implements many aspects of a global workspace via mechanisms of self attention and cross attention.[27] So there is already a robust research program addressing what is in effect a seventh challenge, to build LLM+s with a global workspace.

---

[25] *Global workspace theory:* Bernard J. Baars, *A Cognitive Theory of Consciousness.* Cambridge University Press, 1988. Stanislas Dehaene, 2014. *Consciousness and the Brain.* Penguin.

[26] Anirudh Goyal, Aniket Didolkar, Alex Lamb, Kartikeya Badola, Nan Rosemary Ke, Nasim Rahaman, Jonathan Binas, Charles Blundell, Michael Mozer, Yoshua Bengio. Coordination among neural modules through a shared global workspace. arXiv:2103.01197, 2021. See also Yoshua Bengio, The consciousness prior. arXiv:1709.0856, 2017.

*X = Unified Agency*

The final obstacle to consciousness in LLMs, and maybe the deepest, is the issue of unified agency. We all know these language models can take on many personas. As I put it in an article on GPT-3 when it first appeared in 2020, these models are like chameleons that can take the shape of many different agents.[28] They often seem to lack stable goals and beliefs of their own over and above the goal of predicting text. In many ways, they don't behave like unified agents. Many argue that consciousness requires a certain unity. If so, the disunity of LLMs may call their consciousness into question.

Again, there are various replies. First: it's arguable that a large degree of disunity is compatible with conscious. Some people are highly disunified, like people with dissociative identity disorders, but they are still conscious. Second: One might argue that a single large language model can support an ecosystem of multiple agents, depending on context, prompting, and the like.

But to focus on the most constructive reply: it seems that more unified LLMs are possible. One important genre is the *agent model* (or person model or creature model) which attempts to model a single agent. One way to do that, in systems such as Character.AI, is to take a generic LLM and use fine-tuning or prompt engineering using text from one person to help it simulate that agent.

Current agent models are quite limited and still show signs of disunity. But it's presumably possible in principle to train agent models in a deeper way, for example training an LLM+ system from scratch with data from a single individual. Of course this raises difficult ethical issues, especially when real people are involved. But one can also try to model the perception-action cycle of, say, a single mouse. In principle agent models could lead to LLM+ systems that are much more unified than current LLMs. So once again, the objection turns into a challenge: build LLM+s that are unified agent models.

I've now given six candidates for the X that might be required for consciousness and missing in current LLMs. Of course there are other candidates: higher-order representation (representing one's own cognitive processes, which is related to self-models), stimulus-independent processing (thinking without inputs, which is related to recurrent processing), human-level reasoning (witness the many well-known reasoning problems that LLMs exhibit), and more. Furthermore, it's entirely possible that there are unknown X's that are in fact required for consciousness. Still, these six arguably include the most important current obstacles to LLM consciousness.

---

[27] Arthur Juliani, Ryota Kanai, Shuntaro Sasai. The Perceiver Architecture is a Functional Global Workspace. *Proceedings of the Annual Meeting of the Cognitive Science Society*, vol. 44, 2021. https://escholarship.org/uc/item/2g55b9xx. Andrew Jaegle et al. Perceiver IO: A general architecture for structured inputs and outputs. arXiv:2107.14795, 2021.

[28] David J. Chalmers, GPT-3 and General Intelligence. *Daily Nous,* July 30, 2020. https://dailynous.com/2020/07/30/philosophers-gpt-3/#chalmers

Here's my assessment of the obstacles. Some of them rely on highly contentious premises about consciousness, most obviously in the claim that consciousness requires biology and perhaps in the requirement of sensory grounding. Others rely on unobvious premises about LLMs, like the claim that current LLMs lack world-models. Perhaps the strongest objections are those from recurrent processing, global workspace, and unified agency, where it's plausible that current LLMs (or at least paradigmatic LLMs such as the GPT systems) lack the relevant X and it's also reasonably plausible that consciousness requires X.

Still: for all of these objections except perhaps biology, it looks like the objection is temporary rather than permanent. For the other five, there is a research program of developing LLM or LLM+ systems that have the X in question. In most cases, there already exist at least simple systems with these X's, and it seems entirely possible that we'll have robust and sophisticated systems with these X's within the next decade or two. So the case against consciousness in current LLM systems is much stronger than the case against consciousness in future LLM+ systems.

## 4. Conclusions

Where does the overall case for or against LLM consciousness stand?

Where current LLMs such as the GPT systems are concerned: I think none of the reasons for denying consciousness in these systems is conclusive, but collectively they add up. We can assign some extremely rough numbers for illustrative purposes. On mainstream assumptions, it wouldn't be unreasonable to hold that there's at least a one-in-three chance—that is, to have a subjective probability or credence of at least one-third—that biology is required for consciousness. The same goes for the requirements of sensory grounding, self models, recurrent processing, global workspace, and unified agency.[1] If these six factors were independent, it would follow that there's less than a one-in-ten chance that a system lacking all six, like a current paradigmatic LLM, would be conscious. Of course the factors are not independent, which drives the figure somewhat higher. On the other hand, the figure may be driven lower by other potential requirements X that we have not considered.

Taking all that into account might leave us with confidence somewhere under 10 percent in current LLM consciousness. You shouldn't take the numbers too seriously (that would be specious precision), but the general moral is that given mainstream assumptions about consciousness, it's reasonable to have a low credence that current paradigmatic LLMs such as the GPT systems are conscious.[29]

---

[29] Compared to mainstream views in the science of consciousness, my own views lean somewhat more to consciousness being widespread. So I'd give somewhat lower credences to the various substantial requirements for consciousness I've outlined here, and somewhat higher credences in current LLM consciousness and future LLM+ consciousness as a result.

Where future LLMs and their extensions are concerned, things look quite different. It seems entirely possible that within the next decade, we'll have robust systems with senses, embodiment, world models and self-models, recurrent processing, global workspace, and unified goals. (A multimodal system like Perceiver IO already arguably has senses, embodiment, a global workspace, and a form of recurrence, with the most obvious challenges for it being world-models, self-models, and unified agency.). I think it wouldn't be unreasonable to have a credence over 50 percent that we'll have sophisticated LLM+ systems (that is, LLM+ systems with behavior that seems comparable to that of animals that we take to be conscious) with all of these properties within a decade. It also wouldn't be unreasonable to have at least a 50 percent credence that if we develop sophisticated systems with all of these properties, they will be conscious. Those figures together would leave us with a credence of 25 percent or more. Again, you shouldn't take the exact numbers too seriously, but this reasoning suggests that on mainstream assumptions, it's reasonable to have a significant credence that we'll have conscious LLM+s within a decade.[30]

---

[30] The philosopher Jonathan Birch ("The Search for Invertebrate Consciousness", *Nous* 56:133-53, 2021) distinguishes approaches to animal consciousness that are "theory-heavy" (assume a complete theory), "theory-neutral" (proceed without theoretical assumptions), and "theory-light" (proceed with weak theoretical assumptions). One can likewise take theory-heavy, theory-neutral, and theory-light approaches to AI consciousness. The approach to artificial consciousness that I have taken here is distinct from these three. It might be considered a *theory-balanced* approach, one that takes into account the predictions of multiple theories, perhaps balancing one's credences according to evidence for those theories or according to acceptance of those theories.

One more precise form of the theory-balanced approach might use data about how widely accepted various theories are among experts to provide credences for those theories, and use those credences along with the various theories' predictions to estimate probabilities for AI (or animal) consciousness. In a recent survey of researchers in the science of consciousness (Jolien C. Frankel et al, "An academic survey on theoretical foundations, common assumptions and the current state of consciousness science", *Neuroscience of Consciousness,* issue 1, 2022), just over 50% of respondents indicated that they accept or find promising the global workspace theory of consciousness, while just under 50% indicated that they accept or find promising the local recurrence theory (which requires recurrent processing for consciousness). Figures for other theories include just over 50% for predictive processing theories (which do not make clear predictions for AI consciousness) and for higher-order theories (which require self-models for consciousness), and just under 50% for integrated information theory (which ascribes consciousness to many simple systems but requires recurrent processing for consciousness). Of course turning these figures into collective credences requires further work (e.g. in converting "accept" and "find promising" into credences), as does applying these credences along with theoretical predictions to derive collective credences about AI consciousness. Still, it seems not unreasonable to assign a collective credence above one in three for each of global workspace, recurrent processing, and self models as requirements for consciousness..

What about biology as a requirement? A 2020 survey of professional philosophers (David Bourget and David Chalmers, "Philosophers on Philosophy: The 2020 PhilPapers Survey", *Philosophers' Imprint,* 2023), around 3% accepted or leaned toward the view that current AI systems are conscious, with 82% rejecting or leaning against the view and 10% neutral. Around 39% accepted or leaned toward the view that future AI systems will be conscious, with 27% rejecting or leaning against the view and 29% neutral. (Around 5% rejected the questions in various ways, e.g. saying that there is no fact of the matter or that the question is too unclear to answer). The future-AI figures might tend to suggest a collective credence of at least one in three that biology is required for consciousness (albeit among philosophers rather than consciousness researchers). The two surveys have less information about unified agency and about sensory grounding as requirements for consciousness.

One way to approach this is via the "NeuroAI" challenge of matching the capacities of various non-human animals in virtually embodied systems.[31] It's arguable that even if we don't reach human-level cognitive capacities in the next decade, we have a serious chance of reaching mouse-level capacities in an embodied system with world-models, recurrent processing, unified goals, and so on. If we reach that point, there would be a serious chance that those systems are conscious. Multiplying those chances gives us a significant chance of at least mouse-level consciousness with a decade.[32]

We might see this as a ninth challenge: build multimodal models with mouse-level capacities. This would be a stepping stone toward mouse-level consciousness and eventually to human-level consciousness somewhere down the line.

Of course there's a lot we don't understand here. One major gap in our understanding is that we don't understand consciousness. That's a hard problem, as they say. This yields a tenth challenge: develop better scientific and philosophical theories of consciousness. These theories have come a long way in the last few decades, but much more work is needed.

Another major gap is that we don't really understand what's going on in these large language models. The project of interpreting machine learning systems has come a long way, but it also has a very long way to go. Interpretability yields an eleventh challenge: understand what's going on inside LLMs.

I summarize the challenges here, with four foundational challenges followed by seven engineering-oriented challenges, and a twelfth challenge in the form of a question.

1. Evidence: Develop benchmarks for consciousness.
2. Theory: Develop better scientific and philosophical theories of consciousness.
3. Interpretability: Understand what's happening inside an LLM.
4. Ethics: Should we build conscious AI?

5. Build rich perception-language-action models in virtual worlds.
6. Build LLM+s with robust world models and self models.
7. Build LLM+s with genuine memory and genuine recurrence.
8. Build LLM+s with a global workspace.
9. Build LLM+s that are unified agent models.
10. Build LLM+s that describe non-trained features of consciousness.
11. Build LLM+s with mouse-level capacities.

12. If that's not enough for conscious AI: What's missing?

---

[31] Anthony Zador et al. Toward next-generation artificial intelligence: Catalyzing the NeuroAI revolution. arXiv:2210.08340, 2022.

[32] At NeurIPS I said "fish-level capacities". I've changed this to "mouse-level capacities" (probably a harder challenge in principle), in part because more people are confident that mice are conscious than that fish are conscious, and in part because there is so much more work on mouse cognition than fish cognition.

On the twelfth challenge: Suppose that in the next decade or two, we meet all the engineering challenges in a single system. Will we then have a conscious AI systems? Not everyone will agree that we do. But if someone disagrees, we can ask once again: what is the X that is missing? And could that X be built into an AI system?

My conclusion is that within the next decade, even if we don't have human-level artificial general intelligence, we may well have systems that are serious candidates for consciousness. There are many challenges on the path to consciousness in machine learning systems, but meeting those challenges yields a possible research program toward conscious AI.

I'll finish by reiterating the ethical challenge.[33] I'm not asserting that we should pursue this research program. If you think conscious AI is desirable, the program can serve as a sort of roadmap for getting there. If you think conscious AI is something to avoid, then the program can highlight paths that are best avoided. I'd be especially cautious about creating agent models. That said, I think it's likely that researchers will pursue many of the elements of this research program, whether or not they think of this as pursuing AI consciousness. It could be a disaster to stumble upon AI consciousness unknowingly and unreflectively. So I hope that making these possible paths explicit at least helps us to think about conscious AI reflectively and to handle these issues with care.

*Afterword*

How do things look now, in July 2023, eight months after I gave this lecture at the NeurIPS conference in late November 2022? While new systems such as GPT-4 still have many flaws, they are a significant advance along some of the dimensions discussed in this article. They certainly display more sophisticated conversational abilities. Where I said that GPT-3's performance often seemed on a par with a sophisticated child, GPT-4's performance often (not always) seems on a par with an knowledgeable young adult. There have also been advances in multimodal processing and in agent modeling, and to a lesser extent on the other dimensions that I have discussed. I don't think these advances change my analysis in any fundamental way, but insofar as progress has been faster than expected, it is reasonable to shorten expected timelines. If that is right, my predictions toward the end of this article might even be somewhat conservative.

---

[33] This final paragraph is an addition to what I presented at the NeurIPS conference.